\pdfoutput=1
\documentclass[10pt,conference]{IEEEtran}
\IEEEoverridecommandlockouts

\usepackage{cite}
\usepackage{amsmath,amssymb,amsfonts}
\usepackage{algorithm}
\usepackage{algpseudocode}
\usepackage{graphicx}
\usepackage{booktabs}
\usepackage{multirow}
\usepackage{xcolor}
\usepackage{url}
\usepackage[hidelinks]{hyperref}
\hypersetup{pdftitle={SAGE: Symbolic Action-Gating and Editing for LLM Task Planners},pdfauthor={Trung Minh Bui, JongSul Moon, YoungOuk Kim, Quang-Ngoc Phung, Se-Woong Jun, Dongin Shin}}
\begin{document}

\title{SAGE: Symbolic Action-Gating and Editing\\
for LLM Task Planners}

\author{Trung Minh Bui, JongSul Moon, YoungOuk Kim, Quang-Ngoc Phung,
Se-Woong Jun, and Dongin Shin$^{*}$%
\thanks{This work has been submitted to the IEEE for possible publication.
Copyright may be transferred without notice, after which this version may no
longer be accessible.}%
\thanks{This work was supported by the Ministry of Trade, Industry and Resources
(MOTIR, Korea), Korea, under the Strategic Technology Development Program
supervised by the Korea Institute for Advancement of Technology (KIAT)
[Grant No.\ P0028443], under the Industrial Technology Innovation Program
(No.\ RS-2023-00232141), by the Institute of Information \& Communications
Technology Planning \& Evaluation (IITP) grant funded by the Korean government
(MSIT) (No.\ RS-2025-25441838), and by the Korea Electronics Technology
Institute (KETI) through its Fundamental Research Support Program.}%
\thanks{The authors are with the Intelligent Robotics Research Center, Korea Electronics Technology
Institute (KETI), Seongnam 13449, Republic of Korea
(e-mails: \{minhtrung, moonjongsul, kimyo, ngocpq, daniel, di\_shin\}@keti.re.kr).}%
\thanks{$^{*}$Corresponding author: Dongin Shin.}}

\maketitle

\begin{abstract}
Large language models (LLMs) are now the default cognitive core of embodied
household agents, yet the plans they emit are rarely checked against a grounded
model of the environment before execution, and the task-success they report is
often measured on benchmarks so saturated that no method can be separated from
another. We present \textbf{SAGE} (Symbolic Action-Gating and Editing),
a single-LLM planner built from two lightweight mechanisms: a
\emph{domain-agnostic} symbolic
\emph{gate} ($\sim$250 lines of Python, zero tokens, $O(|\pi|)$) that blocks
precondition-violating actions with \emph{typed} reasons as a runtime safety
monitor, and a local \emph{edit} that regenerates only the failed sub-goal's
suffix, keeping completed and untouched work intact; a hybrid seed$+$live memory
store supports cold-start coverage. Under a leak-free protocol (leave-one-out
retrieval) over five open-weight models and a 75-task AI2-THOR benchmark, we
report the full picture. On the standard benchmark
goal-completeness \emph{saturates} (52\% of instances trivially solved) and SAGE
ties strong hierarchical baselines; we therefore stress the regime where methods
differ. On a harder, method-agnostic multi-goal composition, SAGE's completeness
lead re-emerges large ($+0.06$ to $+0.23$ across four models). Under injected
mid-execution failures, SAGE
recovers as reliably as whole-plan replanners at $2.4$--$3.3\times$ fewer LLM
calls. As a verify-before-execute gate, the symbolic monitor blocks unsafe actions
before actuation and raises \emph{simulator-reported} step-success for every
planner tested (up to $+0.11$), a signal the verifier never sees, so the gain is
non-circular. Because the gate calls no model ($0.008$\,ms/plan), it is a safety
layer that runs essentially free on the edge: SAGE planning reproduces its quality
on a Jetson AGX Orin, where small-model verification helps most. We release the
benchmark, the leak-free protocol, the recovery and safety-gate harnesses, and a
verifier-portability study (auto-induced on ALFWorld, $0.89$ held-out).
\end{abstract}


\section{Introduction}\label{sec:intro}
\begin{figure}[t]
  \centering
  \includegraphics[width=50ex, height=45ex]{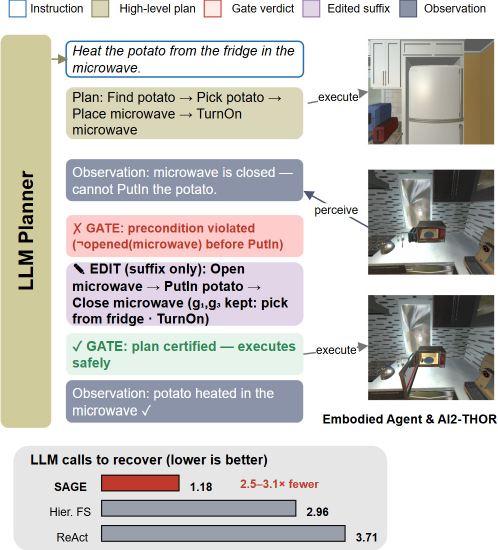}
  \caption{\textbf{SAGE gates and edits.} A naive plan Picks the potato before
  the fridge is open; the zero-token symbolic \emph{gate} flags the precondition
  violation and the sub-goal--local \emph{edit} regenerates only the failed
  sub-goal's suffix, yielding a certified plan without a whole-plan replan
  (frames: AI2-THOR). \emph{Bottom:} under injected mid-execution failures (all
  methods recover $100\%$), SAGE recovers with $2.4$--$3.1\times$ fewer LLM calls
  than whole-plan replanning (up to $3.3\times$ on compound tasks).}
  \label{fig:impact}
\end{figure}

Large language models (LLMs) are increasingly the cognitive
core of embodied agents: they read a goal in natural language, an
action list, and emit a plan a low-level
controller executes~\cite{ahn2022saycan, huang2022inner,
singh2022progprompt, liang2022code}. Even modest open-weight LLMs achieve high \emph{task
success} on benchmarks like ALFWorld~\cite{shridhar2021alfworld} and
AI2-THOR~\cite{kolve2020thor}, but at a hidden cost. Plans average $2$--$3\times$
more steps than the ground-truth reference\footnote{Measured on
pyplanner's 38-task AI2-THOR set;
see \texttt{record\_report.json} in the artifact.}, and execution
``succeeds'' only because most agents add a generous retry-on-failure
loop that masks brittle plans behind extra LLM calls, each paid on a robot's
limited on-board compute, where a \emph{zero}-model-cost grounding layer
(real time on a Jetson AGX Orin) matters most.
\begin{figure*}[t]
  \centering
  \includegraphics[width=105ex, height=50ex]{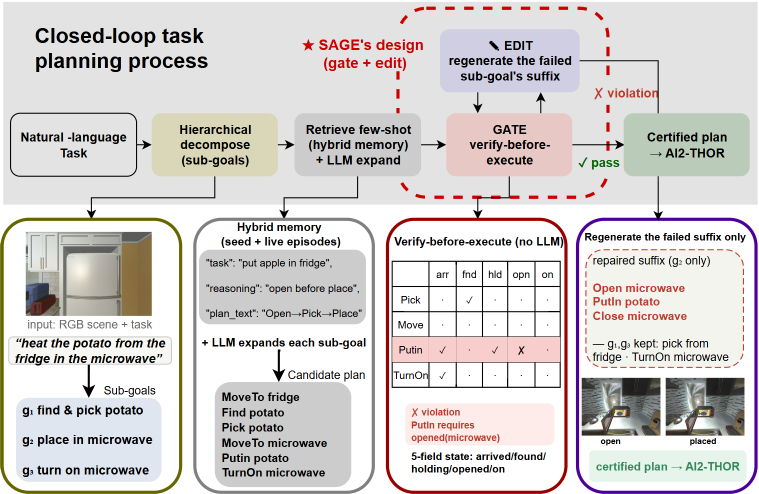}
  \caption{\textbf{SAGE pipeline.} A task is hierarchically decomposed; each sub-goal is expanded with few-shot retrieval from hybrid memory, then every action passes a zero-token symbolic \emph{gate} (verify-before-execute). A flagged precondition violation triggers a sub-goal-local \emph{edit} that regenerates only the failed sub-goal's suffix and re-enters the gate; gate-passing steps form the certified plan executed in AI2-THOR.}
  \label{fig:pipeline}
\end{figure*}

Two design choices in current planners, plus one gap in how we
\emph{evaluate} them, are responsible. \textbf{(P1)} Planners do
not verify plans against a grounded environment model before
execution: preconditions like \emph{``Find an object before
Picking it''} live only in the LLM's prompt and are enforced only
when the simulator errors. This grounded check is \emph{domain-agnostic}
and cheap: a compact rule set ($\sim$250 lines) from the action schema
(STRIPS-style preconditions/effects), needing no expert and auto-inducible
from data (\S\ref{sec:generalization}). \textbf{(P2)} On
failure, planners typically replan the entire remaining task, paying the full
token cost again for context that has already been handled correctly.
\textbf{(P3)} \emph{Methodologically}, standard benchmarks are so saturated
they can no longer separate strong methods. All three are tractable
without retraining the LLM.

We propose \textbf{SAGE} (Symbolic Action-Gating and Editing),
a single-LLM planner whose name reflects its two mechanisms: a symbolic
\emph{gate} on actions and a local \emph{edit} of plans. It integrates
three components:
\begin{enumerate}
  \item a \textbf{rule-based symbolic precondition verifier}
        that runs in $O(|\pi|)$ time and zero tokens and \emph{gates}
        precondition-violating actions before execution,
  \item \textbf{hierarchical decomposition with sub-goal--localised
        editing}: when a step fails, only the suffix of the failed
        sub-goal is regenerated, and
  \item a supporting \textbf{hybrid memory store} that combines a curated
        seed pool of ground-truth plans with a growing pool of
        successful runtime episodes for cold-start coverage.
\end{enumerate}

Building on a preliminary plan-quality analysis~\cite{grace2026ijcas}, this
paper develops a runtime gate-and-edit study; our contributions are
\emph{empirical} and include negative results:

\begin{itemize}
  \item \textbf{The symbolic gate is a non-circular runtime safety
    monitor (addresses P1).} As a verify-before-execute gate, the zero-token
    verifier raises \emph{simulator-reported} step-success for every planner
    (up to $+0.11$ for an unguarded Direct planner; smallest, $+0.02$, for
    SAGE, which already verifies internally) --- measured by an execution signal
    the verifier never observes, so the gain is independent of the
    verifier's own score (\S\ref{sec:sim}).
  \item \textbf{The sub-goal-local edit recovers failures cheaply
    (addresses P2).} Under injected mid-execution failures, SAGE's suffix-only
    edit recovers as reliably as whole-plan replanning (Self-Refine, ReAct,
    Hierarchical Few-Shot) at $2.4$--$3.3\times$ fewer LLM calls: it
    regenerates only the failed sub-goal's suffix and re-certifies it with the
    zero-token verifier, avoiding the re-explanation of already-correct context
    (\S\ref{sec:sim}, \S\ref{sec:ablation}).
  \item \textbf{We expose benchmark saturation and re-separate methods
    (methodological).} SAGE does not \emph{fix} saturation; we make it
    measurable. On the standard benchmark
    goal-completeness saturates ($52\%$ of instances solved by every strong
    baseline) and SAGE ties; on a method-agnostic harder benchmark
    (multi-goal task composition) SAGE's lead re-emerges
    large ($+0.06$ to $+0.23$ completeness points over the strongest
    baseline, Hierarchical Few-Shot, across four models), with
    shorter, more valid plans (\S\ref{sec:compound}). The \textbf{hybrid memory}
    is a supporting component, not a core solver: a curated seed pool
    gives cold-start coverage a live-only pool lacks.
  \item \textbf{A leak-free protocol and reproducible
    artifacts.}\footnote{Code, the benchmark, the human-evaluation bundle, and an
    online appendix (\texttt{online\_appendix.md}, all supplementary tables)
    are available at \url{https://github.com/mtbui2010/sage_release}.} A
    leave-one-out retrieval protocol removes seed$\leftrightarrow$test leakage;
    we demote the verifier-computed \texttt{precondition\_strict} to a
    diagnostic and report metric-validity failures (a reward-based proxy
    ceiling; a goal-checker scoring $8\%$ on reference plans) as findings.
    We release the benchmark, generators, the recovery/verify-gate harnesses,
    and the portability study (rules auto-induced on ALFWorld, $0.89$ held-out).
\end{itemize}

\section{Related Work}\label{sec:related}

\paragraph*{LLM-as-planner.}
SayCan~\cite{ahn2022saycan} grounds skill selection in a learned
affordance value; Inner Monologue~\cite{huang2022inner} adds
textual feedback per step.
ProgPrompt~\cite{singh2022progprompt} and Code-as-Policies~\cite{liang2022code}
push the LLM toward program-style output. LLM+P~\cite{liu2023llmp}
translates NL to PDDL and delegates to a classical planner;
LLM-DP~\cite{dagan2023llm} maintains a symbolic world model
alongside the LLM; recent surveys~\cite{jang2024foundation, bian2025llm}
catalogue these designs through 2024--2025. SAGE differs
by keeping the planner LLM-driven but inserting a deterministic
check between plan and execution.

\paragraph*{Reasoning, refinement, and self-critique.}
Chain-of-Thought~\cite{wei2022chain} elicits intermediate reasoning;
ReAct~\cite{yao2022react} interleaves thought and action;
Tree-of-Thoughts~\cite{yao2023tree} searches over reasoning branches.
Self-Refine~\cite{madaan2023self} and Reflexion~\cite{shinn2023reflexion}
critique with the same LLM. Concurrent 2025 work such as
S\textsuperscript{2}R~\cite{ma2025self} trains the self-critique
via reinforcement learning, keeping the critic inside
the LLM. SAGE instead replaces the LLM critic with a zero-cost
symbolic checker emitting typed feedback, sidestepping the
LLM self-evaluation bias.

\paragraph*{Embodied planning with verification.}
LLM-Planner~\cite{song2023llm} retrieves few-shot examples by
embedding similarity. AdaPlanner~\cite{sun2023adaplanner} distinguishes
in-plan refinement from out-of-plan replanning.
ISR-LLM~\cite{zhou2024isr} validates LLM-emitted PDDL with a
classical solver. LLM-Modulo~\cite{icml2024kambhampati}
generalises ``LLM-generate + external-verifier'';
PDDL-Instruct~\cite{verma2025pddl} instruction-tunes an LLM
to emit PDDL with logical chain-of-thought, reporting up to
94\,\% planning accuracy on classical domains. Parallel safety work
in mobile robotics~\cite{hwang2024safe} applies safety
filtering (control barrier functions, reference governors) at the
\emph{motion} layer; SAGE applies an analogous check-before-execute
discipline at the \emph{plan} layer with a \emph{hand-written,
vocabulary-specific} verifier costing orders of magnitude less
than a classical-planner verifier.

\paragraph*{Memory-augmented agents.}
Voyager~\cite{wang2023voyager} stores reusable code skills;
ExpeL~\cite{zhao2024expel} distills insights from
past trajectories; AutoGuide~\cite{fu2024autoguide} produces
state-conditioned guidelines. SAGE's memory is
\emph{hybrid}: a curated seed pool gives cold-start coverage while a
live pool accumulates successful plans, both surfaced via
the same prompt-block format.

\paragraph*{Hierarchical and repair-oriented planning.}
HiP~\cite{ajay2023hip} and the pyplanner Hierarchical and
HierarchicalFewShot strategies decompose tasks into sub-goals.
DEPS~\cite{wang2023deps} adds failure description + sub-goal
selection in Minecraft. ReplanVLM~\cite{mei2024replanvlm} uses a VLM
to detect failures and trigger replanning. To our knowledge SAGE is
the first to (a) retain sub-goal boundaries \emph{at execution time}
and (b) restrict repair to the failed sub-goal's suffix, preserving
already-completed work and untouched future sub-goals.

\paragraph*{Positioning.}
Verification-based planners typically require a PDDL
domain and an external classical solver
(LLM+P, ISR-LLM)~\cite{liu2023llmp,zhou2024isr} or keep the critic
inside the LLM (Self-Refine, Reflexion)~\cite{madaan2023self,shinn2023reflexion};
SAGE's hand-written, vocabulary-specific verifier needs
neither and costs zero tokens. Repair-oriented planners regenerate the
whole remaining plan; SAGE repairs only the failed sub-goal's suffix.
While many systems are evaluated on offline plan metrics, SAGE
is grounded in simulated \emph{task-success} execution. Finally, a
preliminary version~\cite{grace2026ijcas} is an offline
plan-quality study; this paper contributes
the runtime verify-before-execute gate, the sub-goal-local edit, and a
leak-free empirical study of both.

\section{Method}\label{sec:method}

We propose \textbf{SAGE}, a planner for embodied LLM agents that
integrates three components into a single inference pipeline:
(i) a lightweight, rule-based symbolic precondition verifier,
(ii) hierarchical decomposition with sub-goal--localised repair, and
(iii) a hybrid retrieval memory seeded with curated ground-truth plans
and grown by successful runtime episodes.

We formalise the problem
(\S\ref{sec:problem}), describe each component
(\S\ref{sec:verifier}--\S\ref{sec:memory}), and present
the planning and repair algorithms
(\S\ref{sec:algorithm}). Fig.~\ref{fig:pipeline} shows the full pipeline.

\subsection{Problem formulation}\label{sec:problem}

A household task is a tuple $\tau = (\ell, \mathcal{S}_0)$ where
$\ell$ is a natural-language description and
$\mathcal{S}_0 = (o_0, V_0)$ is the initial scene context: a textual
observation $o_0$ and a set of visible objects $V_0$. A plan
$\pi = (a_1, \dots, a_T)$ is a sequence of actions
$a_t = (\mathrm{verb}_t, \mathrm{obj}_t)$ drawn from a fixed
vocabulary $\mathcal{A}$ (for example,
\{\texttt{MoveTo}, \texttt{Find}, \texttt{Pick}, \texttt{Place},
\texttt{Open}, \texttt{Close}, \texttt{TurnOn}, \texttt{TurnOff}\}
plus body/idle actions); this is the concrete instantiation we
use, but the formulation is agnostic to its contents.

Each action has a precondition $\mathrm{pre}(a)$ and an effect
$\mathrm{eff}(a)$ over a symbolic state $\sigma$ with five
fields: $\mathrm{arrived}, \mathrm{found}, \mathrm{holding},
\mathrm{opened}, \mathrm{on}$ --- the minimal predicates
needed for the preconditions of $\mathcal{A}$. The representation is
extensible: new action types contribute their own fields. A plan is \emph{symbolically valid}
iff $\mathrm{pre}(a_t)$ holds in
$\sigma_{t-1} = \mathrm{eff}(a_{t-1}) \circ \cdots \circ \mathrm{eff}(a_1)(\sigma_0)$
for every $t$. We let $\mathcal{V}(\pi, \sigma_0) \in \{\text{OK}\}
\cup \mathcal{R}$ denote the verifier verdict, where $\mathcal{R}$ is the
typed set of violation reasons (e.g.,
\textsc{NoFindBeforePick}, \textsc{PlaceWrongLocation}).

The planner must produce a $\pi$ that is symbolically valid and
\emph{achieves} the goal when executed in the AI2-THOR
simulator. We assume access only to an LLM $f_\theta$ and the
verifier $\mathcal{V}$, not a PDDL domain or
classical planner.

\subsection{Symbolic Action-Gating verifier}\label{sec:verifier}
\begin{algorithm}[t]
\caption{\texttt{simulate}: rule-based action-gating verification.}\label{alg:verify}
\begin{algorithmic}[1]
\State $\sigma \gets \sigma_0$,\ $\mathit{viol} \gets [\,]$
\For{$t = 1$ \textbf{to} $|\pi|$}
  \State $(\mathit{ok}, r) \gets \texttt{verifyStep}(a_t, \sigma)$
  \If{$\neg \mathit{ok}$}
     \State $\mathit{viol}.\textsc{append}((t, a_t, r))$;\ \textbf{break}
  \EndIf
  \State $\sigma \gets \texttt{apply}(a_t, \sigma)$
\EndFor
\State \Return $(\mathit{viol}=\emptyset,\ \mathit{viol},\ \sigma)$
\end{algorithmic}
\end{algorithm}

The verifier is SAGE's \emph{Action-Gating} mechanism: it gates
each action against the symbolic precondition state before execution,
a \emph{verify-before-execute}
discipline admitting only symbolically valid plans into the
simulator. It is a deterministic function over the schema
in \texttt{pyplanner.base.STEP\_SCHEMA}. Critically, the verifier \emph{never}
invokes an LLM: the gate
$\mathcal{V}$ runs in $O(|\pi|)$ time, costs zero tokens, and gives a
\emph{typed} rejection reason rather than the free-form
critique of self-refinement methods.

The rules are \emph{hand-written} ($\sim$250 lines,
LLM-free) from the AI2-THOR action ontology shipped with
\texttt{pyplanner} (the same ontology the simulator-backed reference
plans satisfy), not from any observation of SAGE's behaviour.
They are not ad-hoc cleverness: each rule instantiates the classical
\emph{precondition--effect} action schema of STRIPS/PDDL~\cite{fikes1971strips}.
Authoring them for a new domain is a mechanical pass over its typed action
interface: for each verb $v\in\mathcal{A}$, list the predicates its execution
\emph{requires} ($\mathrm{pre}$) and \emph{toggles} ($\mathrm{eff}$), both
already declared by any discrete action API. The recipe thus transfers beyond
manipulation (e.g.\ navigation, tool use) without reward models or expert tuning.
The hand-authoring is, moreover, not a permanent porting cost and needs
\emph{no domain expert}: the same rules are \emph{auto-inducible} from logged
interaction across three ontologies, and the result is robust to rule quality
(\S\ref{sec:generalization}), so the manual pass is a one-time convenience. The
\texttt{precondition\_strict} metric in
Sec.~\ref{sec:experiments} is therefore an
\emph{environment-grounded} check, not a self-grading
artefact: it does flag violations on SAGE's own
output, and the identical rule set drives the heuristic
\texttt{precondition} column in our tables.

Algorithm~\ref{alg:verify} sketches \texttt{simulate}.
$\mathrm{pre}(\cdot)$ is implemented as a switch over the action's
verb; the most common rules are:
\begin{itemize}
  \item \texttt{Pick} requires $\mathrm{found} \neq \emptyset$
    and $\mathrm{holding} = \emptyset$.
  \item \texttt{Place~$r$} requires
    $\mathrm{holding} \neq \emptyset \wedge \mathrm{arrived} = r$.
  \item \texttt{Open}/\texttt{Close}/\texttt{TurnOn}/\texttt{TurnOff} require
    $\mathrm{found} = $ the action's object.
  \item Container rule: \texttt{PutIn~$c$} requires
    $c \in \mathrm{opened}$.
\end{itemize}
Violations are formatted into a compact natural-language
feedback block by
\texttt{format\_\allowbreak{}violations\_\allowbreak{}for\_\allowbreak{}llm};
\S\ref{sec:experiments} shows this typed feedback drives more reliable
single-pass refinement than the free-form critiques of
Self-Refine~\cite{madaan2023self}.

\begin{algorithm}[t]
\caption{\textsc{Plan} — SAGE initial planning.}\label{alg:generate}
\begin{algorithmic}[1]
\Require task $\ell$, observation $o$, visible objects $V$
\State $E \gets \textsc{Retrieve}(\mathcal{M}, \ell, k)$
\State $g_{1:K} \gets f_\theta^{\text{decompose}}(\ell, E)$
\State $\sigma \gets (\emptyset, \emptyset, \emptyset, \emptyset, \emptyset),\ V$
\State $B \gets [\,]$
\For{$k = 1$ \textbf{to} $K$}
  \State $\pi^{(k)} \gets f_\theta^{\text{expand}}(g_k, \sigma, E)$
  \For{$j = 1$ \textbf{to} $\texttt{max\_refines}$}
    \State $(\mathit{ok},\mathit{viol}) \gets \mathcal{V}(\pi^{(k)}, \sigma)$
    \If{$\mathit{ok}$}\ \textbf{break}\ \EndIf
    \State $\pi^{(k)} \gets f_\theta^{\text{refine}}(g_k, \pi^{(k)}, \mathit{viol})$
  \EndFor
  \State $\sigma \gets \texttt{rollForward}(\pi^{(k)}, \sigma)$;\
        $B.\textsc{append}((g_k, \pi^{(k)}))$
\EndFor
\State \Return $\textsc{Flatten}(B)$
\end{algorithmic}
\end{algorithm}

\subsection{Hierarchical decomposition with sub-goal blocks}\label{sec:hierarchy}

SAGE decomposes the task into 2--5 sub-goals
$g_1, \dots, g_K$ via one high-level LLM call, then
expands each by a second-level call into a contiguous block of
steps $\pi^{(k)}$. We retain the sub-goal boundaries explicitly as
typed \texttt{\_SubgoalBlock} objects; the flat plan
$\pi = \pi^{(1)} \mathbin{\Vert} \cdots \mathbin{\Vert} \pi^{(K)}$ is
produced only at execution time. SAGE's four LLM roles use fixed system
prompts, given verbatim in the online appendix.

The boundaries enable sub-goal--localised repair
(\S\ref{sec:algorithm}), SAGE's \emph{Editing} mechanism: on
failure it edits only the failed sub-goal's suffix, preserving
already-completed steps and untouched future blocks. To
our knowledge this is the
first work to combine hierarchical LLM planning with
\emph{suffix-only} repair scoped to the failed sub-goal; prior
hierarchical methods (e.g.~\cite{ajay2023hip},
HierarchicalFewShot in pyplanner) replan the entire remainder
on failure.

\subsection{Hybrid memory retrieval}\label{sec:memory}

SAGE conditions both decomposition and expansion on $k$ retrieved
exemplars from a hybrid store $\mathcal{M} = \mathcal{M}_{\text{seed}}
\cup \mathcal{M}_{\text{live}}$:

\begin{itemize}
  \item $\mathcal{M}_{\text{seed}}$ is loaded once from a curated
    ground-truth file (38 simulator-verified plans), guaranteeing
    cold-start coverage before any episode is
    logged. Action names are normalised to the canonical vocabulary
    via \texttt{normalize\_plan} before storage.
  \item $\mathcal{M}_{\text{live}}$ is appended on each successful
    task completion. Persistence is a JSONL file;
    embedding search is optional via the Chroma vector store, with a
    Jaccard-similarity fallback so the system runs on a minimal
    install.
\end{itemize}

Retrieval returns the top-$k$ exemplars by similarity to $\ell$.
Each exemplar contributes a typed
$(\mathrm{task}, \mathrm{reasoning}, \mathrm{plan})$ triple
rendered into a structured
\texttt{=== RETRIEVED EXAMPLE === } block, the same shape used by
pyplanner's \texttt{HierarchicalFewShotPlanner}, so the prompt
engineering is shared and only the store \emph{contents}
differ.

\subsection{Algorithms}\label{sec:algorithm}

Algorithms~\ref{alg:generate} and~\ref{alg:replan} present
$\textsc{Plan}$ and $\textsc{Replan}$ in full.
$\textsc{Plan}$ costs $1 + K + R$ LLM calls, with $K$ sub-goals
and $R \leq K$ refinement passes
from verifier rejection. $\textsc{Replan}$ costs at most two
LLM calls and \emph{never} replans already-executed steps.

\begin{algorithm}[t]
\caption{\textsc{Replan} — sub-goal--localised suffix editing.}\label{alg:replan}
\begin{algorithmic}[1]
\Require completed $C$, failed step $a^\star$, reason $r$, blocks $B$ (Alg.~\ref{alg:generate})
\State $\sigma^\star \gets \texttt{simulate}(C).\mathrm{final}$
       \Comment{recover state at failure, no LLM}
\State $(g^\star, B^{>}) \gets \texttt{locate}(B, |C|)$
       \Comment{which sub-goal failed; what's still in the future}
\State $E \gets \textsc{Retrieve}(\mathcal{M}, g^\star, k)$
\State $\pi^{\text{suf}} \gets f_\theta^{\text{repair}}(g^\star,
       \sigma^\star, a^\star, r, E)$
\State $(\mathit{ok}, \mathit{viol}) \gets \mathcal{V}(\pi^{\text{suf}}, \sigma^\star)$
\If{$\neg \mathit{ok}$}
  \State $\pi^{\text{suf}} \gets f_\theta^{\text{refine}}(g^\star,
         \pi^{\text{suf}}, \mathit{viol})$
\EndIf
\State \Return $\pi^{\text{suf}} \mathbin{\Vert} \textsc{Flatten}(B^{>})$
\end{algorithmic}
\end{algorithm}

\paragraph*{What is novel.}
Each component has precedent: symbolic verification (LLM+P~\cite{liu2023llmp},
ISR-LLM~\cite{zhou2024isr}), hierarchical decomposition (HiP~\cite{ajay2023hip}),
and memory-augmented retrieval (LLM-Planner~\cite{song2023llm},
ExpeL~\cite{zhao2024expel}). SAGE's contribution is their integration under three
constraints: (i) the verifier is hand-written and \emph{LLM-free}, costing zero
tokens; (ii) repair is \emph{strictly local} to the failed sub-goal, preserving
completed and untouched future work; and (iii) the memory is \emph{hybrid},
seeding curated plans and growing from successful runs, with a Jaccard fallback
when embeddings are unavailable.

\section{Experiments}\label{sec:experiments}

We evaluate three claims about SAGE.
\textbf{(C1) Where the benchmark can discriminate, SAGE's plans are better.}
On the standard 75-task benchmark goal-completeness is \emph{saturated}
($52\%$ of instances solved by every strong baseline), so no method
separates and SAGE merely ties --- which we report directly. On a harder,
method-agnostic multi-goal composition, where headroom
returns, SAGE's completeness lead is large and \emph{statistically
significant} (pooled $+0.121$, $p<10^{-4}$; significant on all four models
under Holm--Bonferroni).
\textbf{(C2) SAGE achieves higher step-level execution success than the
baselines.} In the AI2-THOR simulator SAGE reaches $0.71$ vs.\ $0.60$ pooled
step-level success against the one-shot and hierarchical baselines, because the
verifier gate removes precondition-violating steps that abort execution
($17\times$ fewer such failures than Direct). We report
goal-\emph{completeness} (\S\ref{sec:main}) and step-level execution
success; we treat in-simulator goal-\emph{condition} success as unreliable on
this benchmark (an open measurement finding, \S\ref{sec:sim}) and do not
lead with it.
\textbf{(C3) SAGE recovers from failures more cheaply than the baselines.}
Editing only the \emph{failed} sub-goal's suffix lets SAGE recover from injected
mid-execution failures at $2.4$--$3.3\times$ fewer LLM calls than whole-plan
replanning, at equal recovery rate.
A roadmap mapping each claim to its comparison, metric, and supporting tables
is given at the start of the online appendix.

\subsection{Setup}\label{sec:setup}

We ask whether grounding a small open-weight planner with a symbolic verifier,
hierarchical local repair, and memory produces plans that are (i) more
\emph{complete}, (ii) more \emph{executable} in a simulator, and (iii)
\emph{cheaper to repair} after a failure, measured by the metrics defined below.

\paragraph*{Benchmark.}
We use a 75-task AI2-THOR household set: the 38 curated, simulator-verified tasks
shipped with \texttt{pyplanner} plus 37 we generate by instantiating
per-scene templates and AI2-THOR-\emph{verifying} each, keeping only tasks
whose every interaction step the simulator accepts. These 37 span four rooms
(kitchen/living-room/bedroom/bathroom) and three difficulty tiers (14 easy / 16
medium / 7 hard), held out from SAGE's seed memory (no leakage). The
compound set (\S\ref{sec:compound}) is built from these single-goal tasks.

\paragraph*{Models.}
We benchmark five open-weight Ollama models over a $\sim\!10\times$ parameter range --- \texttt{llama3.2}~(3B), \texttt{qwen2.5:7b}, \texttt{mistral-nemo}~(12B), \texttt{qwen2.5:14B} and \texttt{qwen2.5:32b} --- through the same backend, varying only the model identifier to guard against single-model artifacts. The leave-one-out completeness and compound studies use the four 3B--14B models (multi-seed); grounded execution and the safety gate, all five.

\paragraph*{Methods compared.}
We compare SAGE against the seven baselines registered alongside it in
\texttt{pyplanner.REGISTRY}: the one-shot \textbf{Direct} (the strongest
single-call baseline), \texttt{CoT}~\cite{wei2022chain},
\texttt{Few-Shot CoT}~\cite{wei2022chain},
\texttt{Self-Refine}~\cite{madaan2023self}, \texttt{ReAct}~\cite{yao2022react},
\texttt{Hierarchical}, and \texttt{Hierarchical Few-Shot}. SAGE runs with
its refine-overwrite guard on by default; we also report
component ablations (\texttt{SAGE-NoVerifier}, \texttt{SAGE-NoRepair},
\texttt{SAGE-NoMemory}) removing the verifier, local repair, and
memory in turn (\S\ref{sec:ablation}).

\paragraph*{Protocol and statistics.}
For memory methods we use a \emph{leave-one-out} protocol: a test task's own seed
exemplar is held out of memory, removing seed$\leftrightarrow$test leakage
(non-memory baselines are leak-free by construction). The base benchmark
contributes $n{=}37$ held-out single-goal tasks and the compound benchmark
$n{=}74$ multi-goal tasks. The 3B--12B models run with three seeds (averaged
per task); 14B/32B are single-seed scaling points. We report means with $95\%$
bootstrap CIs ($10^4$ resamples) and test headline comparisons with a paired
sign-flip permutation test ($2{\times}10^4$ permutations, $p<0.05$; compound
family Holm--Bonferroni corrected). Full statistics in the online appendix.

\paragraph*{Metrics.}
We report four quantities; arrows give the good direction.
\texttt{completeness}~($\uparrow$) is plan \emph{coverage}: the fraction of the
task's expected interactable objects the plan touches (a goal-coverage proxy
in the spirit of LLM-planning benchmarks~\cite{liu2023llmp}).
\texttt{step\_success}~($\uparrow$), our primary grounded metric, is the fraction
of a plan's steps the AI2-THOR simulator accepts.
\texttt{precondition\_strict}~($\uparrow$) is the fraction of steps whose
preconditions hold against the symbolic state --- a plan-\emph{validity} score.
\texttt{llm\_calls}~($\downarrow$) (with \texttt{tokens}) is the planning/recovery
cost.

\subsection{Main results --- plan quality}\label{sec:main}

Under the leak-free leave-one-out protocol (full per-model grid in the
online appendix), goal completeness on the 75-task benchmark tells a deliberately
unflattering story. \emph{We do not claim a headline completeness win.}
Completeness is \emph{saturated}: across models $52\%$ of
task-instances are already solved (completeness $=1.0$) by every strong
hierarchical baseline, leaving little headroom. SAGE \emph{ties} the strongest
baselines on the base set --- in fact marginally behind Hierarchical Few-Shot on
every model ($0.870$ vs $0.878$ on \texttt{qwen2.5:7b}; $0.836$ vs $0.858$ on
\texttt{mistral-nemo}), all within seed variance. This is a
benchmark property, not a method one: a metric maxed on roughly half the
tasks cannot separate planners. Easy tasks saturate for all
methods while SAGE leads where headroom remains (hard tasks: $0.811$ vs $0.755$
for the strongest hierarchical baseline, leak-free), so
\S\ref{sec:compound} stresses the regime where methods differ. The hard residual
that \emph{no} method solves reflects a grounding gap (object-name
hallucination, long-horizon coverage), not a planning-strategy one. We report
\texttt{precondition\_strict} only as a circular diagnostic (per-model values in
the online appendix).

\subsection{Stressing the saturated metric: harder multi-goal tasks}
\label{sec:compound}

The completeness tie in \S\ref{sec:main} is a ceiling artifact. To recover
headroom we construct a harder benchmark \emph{method-agnostically}:
we compose same-scene single-goal tasks into $2$- and $3$-goal compound tasks,
de-duplicating equivalent compositions
(74 tasks; mean expected-object count $3.7$ vs $2.0$ in the base set); each
reference plan concatenates its already simulator-verified sub-plans.
The construction references nothing about SAGE; it lengthens the horizon
and unions the goal set, forcing a planner to cover \emph{all} sub-goals.

\begin{table}[t]
\centering
\caption{Goal completeness on the harder multi-goal compound benchmark
(leak-free). The base-set completeness tie disappears: SAGE leads by a wide,
consistent margin that \emph{grows} as the base model weakens, while emitting
shorter, more valid plans than the verbose hierarchical baseline.}
\label{tab:compound}
\resizebox{\columnwidth}{!}{%
\begin{tabular}{lcccccc}
\toprule
Model & Direct & Hier-FS & \textbf{SAGE} & $\Delta$ & SAGE steps & SAGE p-valid \\
\midrule
\texttt{qwen2.5:7b} & 0.718 & 0.776 & \textbf{0.854} & \textbf{+0.078} & 12.0 & 0.943 \\
\texttt{qwen2.5:14b} & 0.757 & 0.783 & \textbf{0.902} & \textbf{+0.119} & 12.0 & 0.937 \\
\texttt{llama3.2} & 0.629 & 0.499 & \textbf{0.728} & \textbf{+0.229} & 9.0 & 0.933 \\
\texttt{mistral-nemo} & 0.570 & 0.824 & \textbf{0.880} & \textbf{+0.057} & 12.7 & 0.952 \\
\bottomrule
\end{tabular}
}
\end{table}

Table~\ref{tab:compound} reports the result. Once the ceiling is removed,
SAGE's completeness lead re-emerges, large and consistent across all four
models: $+0.078$ (\texttt{qwen2.5:7b}), $+0.119$ (\texttt{qwen2.5:14b}),
$+0.229$ (\texttt{llama3.2}) and $+0.057$ (\texttt{mistral-nemo}) over the
strongest baseline (Hierarchical Few-Shot). The advantage is one of
\emph{brevity and validity}, not per-step grounding: SAGE reaches higher
coverage with \emph{fewer} steps (Hierarchical Few-Shot averages $17.8$ vs
SAGE's $12.0$ on \texttt{qwen2.5:7b}; Table~\ref{tab:compound}) and
higher precondition validity ($0.94$ vs $0.89$), while its per-step
hallucination rate (fraction of steps whose object is absent from the visible
set) matches the baseline ($0.53$ vs $0.55$). The baseline covers the same
sub-goals only by emitting many more steps; SAGE does so with a shorter,
precondition-valid plan. The advantage
is largest for the weakest model, where naive planners bloat and drift.

\paragraph*{Significance.} The gain over Hierarchical Few-Shot is significant
(paired sign-flip permutation, $20$k permutations): all four models reach
$p<0.05$ and survive Holm--Bonferroni (per-model values in the online appendix).
Pooled over $296$ instances the mean gain is $+0.121$ ($95\%$ bootstrap CI
$[+0.092,+0.150]$, $p<10^{-4}$). Unlike the saturated base benchmark the metric
here discriminates, and the advantage is no leakage artifact: the harder tasks are
a method-agnostic composition under the same leave-one-out protocol.

\subsection{Grounded execution in AI2-THOR}\label{sec:sim}

Offline metrics are proxies, so we execute every plan in AI2-THOR through a ZMQ
action server (up to \texttt{max\_replan}${=}2$ repair rounds), recording per step
whether the simulator accepts the interaction. We run \{\texttt{Direct},
\texttt{Hierarchical}, SAGE\} on all five models over the 38 simulator-verified GT
tasks. These tasks overlap the seed memory, so this comparison favours
memory-using SAGE; the \emph{leak-free} execution evidence is the held-out
safety-gate study (\S\ref{sec:safety}), where the verifier gate, not memory,
drives the gain.

\paragraph*{(O-exec) Grounded plans execute more of their steps.}
Step-level execution success (per-model table in the online appendix): SAGE's plans
reach a higher fraction of accepted steps than both the one-shot and the
hierarchical baseline on \emph{all five} models, lifting the pooled mean
from $0.60$ (Direct) and $0.63$ (Hierarchical) to $0.71$. The mechanism is
direct: SAGE's verifier gate certifies every step's preconditions
before dispatch, removing at plan time the actions the simulator would reject
mid-rollout. The failure taxonomy (online appendix) makes this explicit: precondition-violation failures fall
from $4.5\%$ for Direct to $0.27\%$ for SAGE (a $17\times$ reduction), and
SAGE's in-simulator strict-precondition score is $0.958$ versus $\approx0.86$
for both baselines. The verifier suppresses the failure class it
targets.

\paragraph*{On goal-condition success.}
We do \emph{not} lead with in-simulator goal-\emph{condition} success: it is low
across \emph{all} methods with differences that are not statistically reliable, so
the available checkers cannot separate planners --- an open measurement finding
(\S\ref{sec:limitations}). For an absolute number we instead use a blinded,
method-shuffled \emph{two-rater human evaluation} (one author, one non-author) of a
72-item subset: SAGE leads with end-task success $0.53$ vs.\ $0.40$ (Direct) and
$0.28$ (Hierarchical), at substantial inter-rater agreement (Cohen's
$\kappa{=}0.65$); the zero-token partial-credit checker tracks the human consensus
at Pearson $r{=}0.82$ (online appendix).

\paragraph*{Failure recovery: equal reliability, far lower cost.}
We inject one mid-execution failure per task and measure recovery \emph{cost}
(details in the online appendix). Recovery \emph{rate} saturates ($100\%$ for every
method here), so cost discriminates. On held-out tasks SAGE recovers in $1.18$ LLM
calls versus $2.87$--$3.71$ for whole-plan replanners (Hier.\ Few-Shot,
Self-Refine, ReAct): a $2.4$--$3.1\times$ reduction at equal reliability and lowest
latency, because the edit regenerates only the failed sub-goal's suffix
($+1.7$ to $+2.5$ fewer calls vs.\ every baseline, all $p<10^{-4}$). On compound
tasks the gap widens to $2.6$--$3.3\times$ (up to $3.7\times$ lower latency): a
global $2.4$--$3.3\times$ envelope.

\subsection{Generalization to unseen environments}\label{sec:generalization}

A standing objection to a symbolic-verifier method is that both the 38 curated
scenes and the hand-written rules overfit iTHOR. We address
this with out-of-distribution studies that change \emph{nothing} in
SAGE --- identical prompts, seed memory, and verifier code.

\paragraph*{Unseen procedurally-generated houses.}
On 70 unseen ProcTHOR houses, SAGE retains its strict-precondition lead ($0.985$): plan \emph{validity} transfers to never-seen layouts (full table in the online appendix).

\paragraph*{Portability: the verifier's rules are \emph{inducible} from data.}
The hand-written verifier ($\sim$250 lines) is not arbitrary tuning: a probe-based
pipeline that mines the conjunctive precondition per action from
$(\text{state},\text{action},\text{success})$ transitions (zero manual rules)
recovers it in-domain (iTHOR, $14/16$ rules), on unseen ProcTHOR ($13/16$), and on
text-domain ALFWorld ($0.89$ held-out), so porting the monitor to a new action
distribution needs interaction data, not fresh authoring. The result is also
robust to rule \emph{quality}: random rule deletion degrades the gate smoothly
and the non-expert induced set matches the hand-written one (online appendix).

\paragraph*{Published baselines.}
Against \textbf{LLM+P}~\cite{liu2023llmp} and a \textbf{SayCan}-style decoder~\cite{ahn2022saycan}, both produce \emph{valid but incomplete} plans (completeness $0.61$--$0.77$) whereas SAGE reaches $0.87$--$0.88$ on the same models (online appendix).

\subsection{External validation: EAI VirtualHome}\label{sec:eai}
To rule out that SAGE's gains are an artifact of its own verifier's metrics, we
evaluate on the independent \emph{Embodied Agent Interface} (EAI) VirtualHome
benchmark~\cite{eai2024}, scored by EAI's \emph{own} transition model. Without
changing any SAGE code, we compare verifier-on against \texttt{SAGE-NoVerifier}
over $305$ tasks, three seeds (per-model table in the online appendix). Two effects are
robust. The verifier removes redundant steps on \emph{every} model ($-2.7$ to
$-7.4$ points, largest on the strongest --- third-party evidence for the
brevity-and-validity mechanism of \S\ref{sec:compound}), and lifts executability
for the small open models ($+3.6$/$+3.7$ on 7B/3B) but \emph{saturates} on 14B
($-1.5$). Correctness is unchanged: the gate improves plan \emph{validity}, not
goal grounding.

\subsection{Runtime safety monitor: blocking erroneous actions before actuation}
\label{sec:safety}
An LLM planner that emits a precondition-violating step --- Pick before the
object is localized, Place into a closed receptacle --- will, without a check,
\emph{send that action to the robot}. Because the verifier is LLM-free it can run
\emph{online} as a verify-before-execute guard: before a step is dispatched we
check it against the symbolic state and, if it would violate a precondition,
repair the sub-goal \emph{before} the action is taken rather than
execute-then-recover after the actuator has already moved. This is the safety
contribution: across four models on the $38$ grounded tasks the gate
\textbf{blocks $206$ precondition-violating actions before they reach the
actuator} (per-method gate-off/on in the online appendix) --- erroneous commands that
the execute-then-replan baseline would have physically attempted. The benefit on
\emph{simulator-reported} step-success is independent of the verifier and, as a
safety layer should, concentrates on the planners that emit the most unsafe steps:
for the unguarded \texttt{Direct} planner the gate raises step-success by
$+0.107$ ($p<10^{-4}$); for planners already embedding structure the effect is
smaller and not significant (\texttt{CoT} $+0.081$, $p{=}0.09$; \texttt{Hier}
$+0.041$; SAGE $+0.016$, both n.s.), as expected --- they emit fewer violations to
catch. We therefore claim the gate as a deployable, zero-token safety layer that
prevents unsafe actuation for \emph{unverified} planners, not a universal accuracy
boost.

\subsection{Ablations and cost}\label{sec:ablation}

Removing any single component degrades SAGE: \texttt{SAGE-NoVerifier}
loses the strict-precondition guarantee, \texttt{SAGE-NoRepair} cannot
recover from injected failures, and \texttt{SAGE-NoMemory} loses cold-start
coverage. We state the cost
directly: SAGE spends roughly $7\times$ the tokens of one-shot Direct,
\emph{comparable} to the most expensive baseline (Hierarchical
Few-Shot), not cheap in absolute terms. Its efficiency claim is narrow:
at \emph{recovery} time, repair scoped to a single sub-goal costs
$2.4$--$3.3\times$ fewer LLM calls than whole-plan replanning --- not a
general token-cost advantage.

\subsection{On-device deployment: where the zero-token gate matters most}
\label{sec:edge}
The safety gate matters most where compute is scarcest. The small ($3$--$7$B)
open-weight models an embodied agent can run \emph{on-board} (when a cloud LLM is
unavailable, too costly, or disallowed) are also the ones that emit the most
precondition-violating actions for the gate to block (\S\ref{sec:safety}) --- and
the verifier, calling \emph{no} model, is the cheap half of that trade. We confirm
this on real edge silicon: an NVIDIA Jetson AGX Orin ($64$\,GB, JetPack~6,
CUDA~12.6, MAXN) serving $4$-bit (Q4\_K\_M) models on-board with Ollama, AI2-THOR
on a host (planner-on-edge, sim-on-host). Three findings (table in the
online appendix). \textbf{(i) The gate is free on-board:} the verifier adds a measured
\textbf{$0.008$\,ms} per plan (median; $0.017$\,ms max, $O(|\pi|)$, no tokens),
six orders of magnitude below the multi-second on-device LLM latency.
\textbf{(ii) No quality loss:} on a balanced $20$-task subset (seed~$0$), on-board
plan quality matches the server within seed variance and SAGE's completeness lift
over an unguarded planner reproduces on-board ($+0.12/+0.16/+0.13$ for
$3$B/$7$B/$14$B). \textbf{(iii) Feasible:} even $14$B runs in $27$\,GB at $48$\,W.
The component that helps small models most is thus essentially free on the edge
--- a deployable, cloud-free safety layer for the on-board planners that need it
most, and a partial mitigation of L1 (decision-making on real edge hardware,
execution still simulated).

\subsection{Limitations}\label{sec:limitations}
\emph{(L1) Simulation only.} All \emph{execution} is in AI2-THOR (no real-robot
claim; an xArm evaluation is future work), though we do run the \emph{planner} on
edge hardware (\S\ref{sec:edge}), a partial mitigation.
\emph{(L2) Saturated base metric.} Base-benchmark completeness saturates ($52\%$
ceiling) and cannot separate methods; we report the tie and locate our gains on the
harder compound benchmark, recovery cost, and the safety gate.
\emph{(L2b) Checker-based end-task success is unreliable.} Both in-simulator
goal-condition checkers fail (online appendix); we therefore obtain the absolute
end-task number from a blinded two-rater human evaluation (72 items,
$\kappa{=}0.65$; online appendix) rather than an automated checker.
\emph{(L3) Hand-written verifier.} The deployed verifier is hand-authored
($\sim$250 lines); porting needs re-authoring or rule induction
(\S\ref{sec:generalization}), not a full PDDL domain.
\emph{(L4) Baseline asymmetries.} SayCan uses our verifier as its affordance
filter (strict-precondition $1.0$ by construction); the ALFWorld result is a
\emph{verifier}-portability study, as the planners' vocabulary (clean/heat/cool)
is not yet adapted there.
\emph{(L5) Residual hallucination.} All methods, SAGE included, still reference
absent objects on some steps --- a shared small-model weakness.
\emph{(L6) Execution subset.} The expensive AI2-THOR execution covers a
representative method subset, not all seven baselines.
\emph{(L7) ALFWorld floored.} On ALFWorld step-executability saturates at $1.0$
while end-task success floors ($\le0.04$): long-horizon ALFRED goals are out of
reach for a 7B planner without fine-tuning, so no method separates.
\emph{(L8) Model-dependent gate benefit.} The gate's executability gain is
regime-specific (\S\ref{sec:eai}): largest for small models, saturating on the
strongest, so we scope the claim to that regime.

\section{Conclusion}\label{sec:conclusion-full}

We presented \textbf{SAGE}, a planner for embodied LLM agents that
couples symbolic action-gating (a rule-based precondition verifier
applied as a verify-before-execute monitor) with sub-goal--localised
editing, supported by hierarchical decomposition and a hybrid memory
store. The components are individually
lightweight: the gate is $\sim$250 lines of deterministic
Python, the edit is restricted to the suffix of a single sub-goal,
and the memory store falls back to a Jaccard retriever when
embeddings are unavailable. Their integration yields a planner whose advantages we report
only where the data supports them. Under a leak-free
leave-one-out protocol over four open-weight models, goal-completeness
on the standard benchmark is saturated and SAGE ties the strongest
baselines. Where the benchmark can discriminate, the advantage is real
and statistically significant: on a harder, method-agnostic multi-goal
benchmark SAGE improves completeness by a pooled $+0.121$
($p<10^{-4}$, significant on all four models under Holm--Bonferroni,
Sec.~\ref{sec:compound}); in grounded AI2-THOR execution SAGE attains
higher \emph{step-level} success ($0.71$ vs.\ $0.60$ pooled,
Sec.~\ref{sec:sim}); we report step- rather than end-task success because the
available goal checkers are unreliable, an open measurement finding. As a
verify-before-execute gate the same check blocks $206$ unsafe actions before
actuation and raises an unverified Direct planner's simulator step-success by
$+0.107$ ($p<10^{-4}$), an independent, non-circular safety result. Because the
gate calls no model ($0.008$\,ms/plan), this safety layer runs essentially free
on-board: SAGE planning reproduces its quality on a Jetson AGX Orin
(Sec.~\ref{sec:edge}), exactly where small-model verification helps most.

On cost, SAGE spends $\sim7\times$ the tokens of a one-shot baseline, comparable
to the most expensive baseline; its efficiency claim is narrow, at \emph{recovery}
time, where sub-goal--local editing costs $2.4$--$3.3\times$ fewer LLM calls than
whole-plan replanning. The dominant residual failure is object-naming ambiguity
(\S\ref{sec:limitations}), orthogonal to planning strategy, motivating future
grounding of the action vocabulary in vision-language embeddings.

\paragraph*{Reproducibility.}
SAGE ships as a self-contained pyplanner strategy
(\texttt{REGISTRY["SAGE"]}) with the verifier, hybrid retriever, and all benchmark
scripts; the simulator-verified $75$-task AI2-THOR set is released with the artifact.

\let\oldthebibliography\thebibliography
\renewcommand{\thebibliography}[1]{\oldthebibliography{#1}\setlength{\itemsep}{-0.8ex}}
\bibliographystyle{IEEEtran}
\bibliography{references}

\end{document}